\documentclass[11pt,a4paper]{article}
\usepackage[hyperref]{emnlp2020}
\usepackage{times}
\usepackage{latexsym}

\usepackage{multicol, multirow}
\usepackage{amsmath, amssymb, amsfonts}
\usepackage{ulem}
\usepackage{graphicx}
\usepackage{arydshln}

\newcommand{\highlight}[3][black]{{\fboxsep0.5pt\colorbox{#2}{\color{#1} #3}}}

\usepackage{microtype}

\aclfinalcopy 
\def\aclpaperid{2712} 

\title{Zero-Shot Stance Detection:\\A Dataset and Model using Generalized Topic Representations}

\author{Emily Allaway \\
  Columbia University \\
  New York, NY \\
  \texttt{eallaway@cs.columbia.edu} \\\And
  Kathleen McKeown \\
  Columbia University \\
  New York, NY \\
  \texttt{kathy@cs.columbia.edu} \\}

\date{}

\begin{document}
\maketitle
\begin{abstract}
Stance detection is an important component of understanding hidden influences in everyday life. Since there are thousands of potential topics to take a stance on, most with little to no training data, we focus on zero-shot stance detection: classifying stance from no training examples. In this paper, we present a new dataset for zero-shot stance detection that captures a wider range of topics and lexical variation than in previous datasets. Additionally, we propose a new model for stance detection that implicitly captures relationships between topics using generalized topic representations and show that this model improves performance on a number of challenging linguistic phenomena. 
\end{abstract}

\section{Introduction}
Stance detection, automatically identifying positions on a specific topic in text \citep{Mohammad2017StanceAS}, is crucial for understanding how information is presented in everyday
life. For example, a news article on crime may also implicitly take a position on immigration (see Table ~\ref{tab:introex}).  

There are two typical approaches to stance detection: \textit{topic-specific} stance (developing topic-specific classifiers, e.g., \citet{Hasan2014WhyAY})  and \textit{cross-target} stance (adapting classifiers from a related topic to a single new topic detection, e.g., \citet{Augenstein2016StanceDW}). Topic-specific stance requires the existence of numerous, well-labeled training examples in order to build a classifier for a new topic, an unrealistic expectation when there are thousands of possible topics 
for which
data collection and annotation 
are both time-consuming and expensive. 
While cross-target stance does not require training examples for a new topic, it
does require human knowledge about any new topic and how it is related to the training topics. 
As a result,
models developed for this variation are still limited in their ability to generalize to a wide variety of topics. 

In this work, we propose two additional variations of stance detection: zero-shot stance detection (a classifier is evaluated on a large number of completely new topics) and few-shot stance detection (a classifier is evaluated on a large number of topics for which it has very few training examples). Neither variation requires any human knowledge about the 
new
topics or their relation to training topics.
Zero-shot stance detection, in particular, is a
more accurate evaluation of a model's ability to generalize
to the range of topics in the real world.

\begin{table}[t]
\begin{tabular}{|ll|}
\hline
\uline{\textbf{Topic:}} immigration & \uline{\textbf{Stance:}} against \\ 
& \\
\multicolumn{2}{|l|}{\begin{tabular}[c]{@{}l@{}}\uline{\textbf{Text:}}  The jury’s verdict will ensure that\\ another \highlight[black]{yellow!30}{violent criminal alien} will be removed\\ from \highlight[black]{yellow!30}{our} \highlight[black]{yellow!30}{community} for a very long period …\end{tabular}} \\ \hline
\end{tabular}
\caption{Example snippet from Fox News describing a crime but taking a stance \textit{against} immigration. Phrases indicating stance are \highlight[black]{yellow!30}{highlighted}.}
\label{tab:introex}
\end{table}
Existing stance datasets typically have a small number of topics (e.g., 6) that are described in only one way (e.g., `gun control'). This is not ideal for zero-shot or few-shot stance detection because there is little linguistic variation
in how topics are expressed (e.g., `anti second amendment')
and limited topics. Therefore, to facilitate evaluation of zero-shot and few-shot stance detection, we create a new dataset, \textbf{VA}ried \textbf{S}tance \textbf{T}opics (VAST). VAST
consists of a large range of topics covering broad themes, such as politics (e.g., `a Palestinian state'), education (e.g., `charter schools'), and public health (e.g., `childhood vaccination'). In addition, the data includes a wide range of similar expressions (e.g., `guns on campus' versus `firearms on campus'). This variation captures how humans might realistically describe the same topic and
contrasts with the lack of variation in existing datasets.

We also develop a model for zero-shot stance detection that exploits information about topic similarity through generalized topic representations obtained through contextualized clustering. These topic representations are unsupervised and therefore represent information about topic relationships without requiring explicit human knowledge. 

Our contributions are as follows: (1) we develop a new dataset, VAST, for zero-shot and few-shot stance detection and 
(2) we propose a new model for stance detection that improves performance on a number of challenging linguistic phenomena (e.g., sarcasm) and relies less on sentiment cues (which often lead to errors in stance classification).
We make our dataset and models available for use: \url{https://github.com/emilyallaway/zero-shot-stance}.

\section{Related Work}
\label{sec:related}
Previous datasets for stance detection have centered on two definitions of the task~\citep{Kk2020StanceD}. In the most common definition (\textit{topic-phrase} stance), stance (pro, con, neutral) of a text is detected towards a topic that is usually a noun-phrase (e.g., `gun control'). In the second definition (\textit{topic-position} stance), stance (agree, disagree, discuss, unrelated) is detected between a text and a topic that is an entire position statement (e.g., `We should disband NATO').

A number of datasets exist using the \textit{topic-phrase} definition with texts from online debate forums \citep{Walker2012ACF,Abbott2016InternetAC,Hasan2014WhyAY}, information platforms \citep{Lin2006WhichSA,Murakami2010SupportOO}, student essays \citep{Faulkner2014AutomatedCO}, news comments \citep{Krejzl2017StanceDI,Lozhnikov2018StancePF} and Twitter \citep{Kk2017StanceDI,Tsakalidis2018NowcastingTS,Taul2017OverviewOT,Mohammad2016SemEval2016T6}. 
These datasets generally have a very small number of topics (e.g., \citet{Abbott2016InternetAC} has 16) and the few with larger numbers of topics \citep{BarHaim2017StanceCO,Gottipati2013LearningTA,Vamvas2020XA} still have limited topic coverage (ranging from $55$ to $194$ topics). The data used by \citet{Gottipati2013LearningTA}, articles and comments from an online debate site, has the potential to cover the widest range of topics, relative to previous work. However, their dataset is not explicitly labeled for topics, does not have clear pro/con labels, and does not exhibit linguistic variation in the topic expressions. 
Furthermore, all of these stance datasets are not used for zero-shot stance detection due to the small number of topics, with the exception of the SemEval2016 Task-6 (TwitterStance) data, which is used for cross-target stance detection with a single unseen topic \citep{Mohammad2016SemEval2016T6}. In constrast to the TwitterStance data, which has only one new topic in the test set, our dataset for zero-shot stance detection has a large number of new topics for both development and testing.

For \textit{topic-position} stance, datasets primarily use text from news articles with headlines as topics \citep{Thorne2018FEVERAL,Ferreira2016EmergentAN}. In a similar vein, \citet{Habernal2018TheAR} use comments from news articles and manually construct position statements. These datasets, however, do not include clear, individuated topics and so we focus on the topic-phrase definition in our work.

Many previous models for stance detection trained an individual classifier for each topic \citep{Lin2006WhichSA,beigman-klebanov-etal-2010-vocabulary, sridhar-etal-2015-joint,Somasundaran2010RecognizingSI,Hasan2013StanceCO,Li2018StructuredRL,Hasan2014WhyAY} or for a small number of topics common to both the training and evaluation sets \citep{Faulkner2014AutomatedCO,Du2017StanceCW}. In addition, a handful of models for the TwitterStance dataset have been designed for cross-target stance detection \citep{Augenstein2016StanceDW,Xu2018CrossTargetSC}, including a number of weakly supervised methods using unlabeled data related to the test topic \citep{Zarrella2016MITREAS,Wei2016pkudblabAS,Dias2016INFUFRGSOPINIONMININGAS}. In contrast, our models are trained jointly for all topics and are evaluated for zero-shot stance detection on a large number of new test
topics (i.e., none of the zero-shot test topics occur in the training data).

\section{VAST Dataset}
We collect a new dataset, VAST, for zero-shot stance detection that includes a large number of specific topics.
Our annotations are done on comments collected from \textit{The New York Times} `Room for Debate' section, part of the Argument Reasoning Comprehension (ARC) Corpus \citep{Habernal2018TheAR}.
Although the ARC corpus provides stance annotations, they follow the \textit{topic-position} definition of stance, 
as in \S\ref{sec:related}. This format makes it difficult to determine stance in the typical \textit{topic-phrase} (pro/con/neutral) setting with respect to a single topic, as opposed to a position statement (see \textit{Topic} and \textit{ARC Stance} columns respectively, Table \ref{tab:dataex}). Therefore, we collect annotations on both topic and stance, using the ARC data as a starting point. 

\begin{table*}[t]
\begin{tabular}{lllll|l}
\textbf{Comment} & \textbf{ARC Stance} & \textbf{Topic} & \textbf{$\mathcal{\ell}$} & \textbf{Type} & \\ \hline

 
\multirow{3}{*}{\begin{tabular}[t]{@{}l@{}} ... Instead they have to work jobs\\ (while \highlight[black]{yellow!30}{their  tax dollars are going to}\\ \highlight[black]{yellow!30}{supporting} illegal aliens) in order to\\put themselves through college {[}cont{]}\end{tabular}} & \multirow{2}{*}{\begin{tabular}[t]{@{}l@{}} \highlight[black]{green!30}{\textit{Immigration}} is \\ really a problem \end{tabular}} & \begin{tabular}[t]{@{}l@{}} immigration \end{tabular}& Con & Heur & \textbf{(1)}\\ \cdashline{3-6}
& & \sout{a problem} \rotatebox[origin=c]{270}{$\Rsh$} &  & \\
& & \begin{tabular}[t]{@{}l@{}}costs to \\ american citizens      \end{tabular} & Con & List & \textbf{(2)} \\ \hline

\multirow{3}{*}{\begin{tabular}[t]{@{}l@{}}Why should it be our job to help out the\\ \highlight[black]{yellow!30}{owners of the  restaurants and bars?} ...\\
If they were paid a \highlight[black]{yellow!30}{living wage} ...[cont]\end{tabular}} & Not to tip & 
\sout{workplace} \rotatebox[origin=c]{270}{$\Rsh$} & & & \textbf{(3)} \\
& & living wage & Pro & Corr & \\ \cdashline{3-6}
& & restaurant owners & Con & List & \textbf{(4)}\\
\hline

\multirow{2}{*}{\begin{tabular}[t]{@{}l@{}}...I like being able to access the internet\\ about my health issues, and find I can \\ talk with my doctors ...  {[}cont{]}\end{tabular}} 
& \begin{tabular}[t]{@{}l@{}}\highlight[black]{green!30}{\textit{\textit{Medical websites}}} \\ are healthful\end{tabular} & medical websites & Pro & Heur & \textbf{(5)}\\ \cdashline{3-6}
 & & home schoolers & Neu & & \textbf{(6)}\\ \hline


\end{tabular}
\caption{Examples from VAST, showing the position statement in the original ARC data and our topics, labels ($\mathcal{\ell}$) and type (see \S3.1). We show extracted topic (\highlight[black]{green!30}{\textit{green, italics}}), extracted but corrected topics (\sout{strikeout}), and phrases that match with annotator-provided topics (\highlight[black]{yellow!30}{yellow}). Neu indicates neutral label.}
\label{tab:dataex}
\end{table*}

\subsection{Data Collection}
\subsubsection{Topic Selection}
To collect stance annotations, we first heuristically extract specific topics from the stance positions provided by the ARC corpus. We define a candidate topic as a noun-phrase in the constituency parse, generated using Spacy\footnote{\url{spacy.io}}, 
of the ARC stance position (as in (1) and (5) Table \ref{tab:dataex}).
To reduce noisy topics, we filter candidates to include only noun-phrases in the subject and object position of the main verb in the sentence. If no candidates remain for a comment after filtering, we select topics from the categories assigned by \textit{The New York Times} to the original article the comment is on (e.g., the categories assigned for (3) in Table \ref{tab:dataex} are `Business', `restaurants', and `workplace'). From these categories, we remove proper nouns as these are over-general topics (e.g., `Caribbean', `Business').
From these heuristics we extract $304$ unique topics from $3365$ unique comments (see examples in Table \ref{tab:dataex}). 

Although we can extract topics heuristically, they are sometimes noisy. For example, in (2) in Table \ref{tab:dataex}, `a problem' is extracted as a topic, despite being overly vague. 
Therefore, we use crowdsourcing to collect stance labels and additional topics from annotators.

\subsubsection{Crowdsourcing}
\label{sec:crowd}
We use Amazon Mechanical Turk to collect 
crowdsourced annotations. We present each worker with a comment and first ask them to list topics related to the comment, to avoid biasing workers toward finding a stance on a topic not relevant to the comment. 
We then provide the worker with the automatically generated topic for the comment and ask for the stance, or, if the topic does not make sense, to correct it. Workers are asked to provide stance on a 5-point scale (see task snapshot in Appendix \ref{app:crowd}) which we map to 3-point pro/con/neutral. Each topic-comment pair is annotated by three workers. We remove work by poor quality annotators, determined by manually examining the topics listed for a comment and using MACE \citep{Hovy2013LearningWT} on the stance labels. For all examples, we select the majority vote as the final label. When annotators correct the provided topic, we take the majority vote of stance labels on corrections to the same new topic. 

Our resulting dataset includes annotations of three types (see Table \ref{tab:dataex}): \textbf{Heur} stance labels on the heuristically extracted topics provided to annotators (see (1) and (5)), \textbf{Corr} labels on corrected topics provided by annotators (see (3)), \textbf{List} labels on the topics listed by annotators as related to the comment (see (2) and (4)). 
We include the noisy type \textbf{List}, because we find that the stance provided by the annotator for the given topic also generally applies to the topics the annotator listed and these provide additional learning signal (see \ref{app:data} for full examples).
We clean the final topics to remove noise by lemmatizing and removing stopwords using NLTK\footnote{\label{foot:nltk}\url{nltk.org}} and running automatic spelling correction\footnote{\url{pypi.org/project/pyspellchecker}}.

\subsubsection{Neutral Examples}
\label{sec:neu}
Every comment will not convey a stance on every topic. Therefore, it is important to be able to detect when the stance is, in fact, neutral or
neither. Since the original ARC data does not include neutral stance, our crowdsourced annotations yield only $350$
neutral examples. Therefore, we add additional examples to the neutral class that are \textit{neither} pro nor con. These examples are constructed automatically by permuting existing topics and comments. 

We convert each entry of type \textbf{Heur} or \textbf{Corr} in the dataset to a neutral example for a different topic with probability $0.5$. We do not convert type noisy \textbf{List} entries into neither examples. If a comment $d_i$ and topic $t_i$ pair is to be converted, we randomly sample a new topic $\Tilde{t_i}$ for the comment from topics in the dataset. To ensure $\Tilde{t_i}$ is semantically distinct from $t_i$, we check that $\Tilde{t_i}$ does not overlap lexically with $t_i$ or any of the topics provided to or by annotators for $d_i$ (see (6) Table \ref{tab:dataex}).

\begin{table}[t]
    \centering
    \begin{tabular}{l|c|c|c}
        \hline
        & \textbf{\#} & \%P & \%C \\ \hline
        \textit{Type Heur} &  4416 & 49 & 51 \\ 
        \textit{Type Corr} &  3594 & 44 & 51\\ 
        \textit{Type List} & 11531 & 50 & 48\\ 
        \textit{Neutral examples} & 3984 & -- & -- \\ \hline
        \textbf{TOTAL examples} & \textbf{23525} & 40 & 41\\ \hline \hline
        \textit{Topics} & 5634 & --& --\\ \hline
    \end{tabular}
    \caption{VAST dataset statistics. P is Pro, C is Con. Example types (\S \ref{sec:crowd}): \textit{Heur} -- original topic, \textit{Corr} -- corrected topic, \textit{List} -- listed topic}
    \label{tab:finaldata}
\end{table}
\subsection{Data Analysis}
\label{sec:datanalysis}
The final statistics of our data are shown in Table \ref{tab:finaldata}. 
We use Krippendorff $\alpha$ to compute interannotator agreement, yielding $0.427$, and percentage agreement ($75\%$), which indicate stronger than random agreement. 
We compute agreement only on the annotated stance labels for the topic provided, since few topic corrections result in identical new topics. We see that
while the task is challenging, annotators agree the majority of the time.

We observe the most common cause of disagreement is annotator inference about stance relative to an overly general or semi-relevant topic. For example, annotators are inclined to select a stance for the provided topic (correcting the topic only $30\%$ of the time),  even when it does not make sense or is too general (e.g., `everyone' is overly general). 

The inferences and corrections by annotators provide a wide range of stance labels for each comment. For example, for a single comment our annotations may include multiple examples, each with different topic and potentially different stance labels, all correct (see (3) and (4) Table \ref{tab:dataex}). That is, our annotations capture semantic and stance complexity in the comments and are not limited to a single topic per text.  This increases the difficulty of predicting and annotating stance for this data. 

In addition to stance complexity, the annotations provide great variety in how topics are expressed, with a median of $4$ unique topics per comment. While many of these are slight variations on the same idea (e.g., `prison privatization' vs. `privatization'), this more accurately captures how humans might discuss a topic, compared to restricting themselves to a single phrase (e.g., `gun control'). The variety of topics per comment makes our dataset challenging and the large number of topics with few examples each 
(the median number of examples per topic is $1$ and the mean is $2.4$) makes our dataset
well suited to developing models for zero-shot and few-shot stance detection.

\begin{figure}[t]
    \centering
    \includegraphics[width=.48\textwidth]{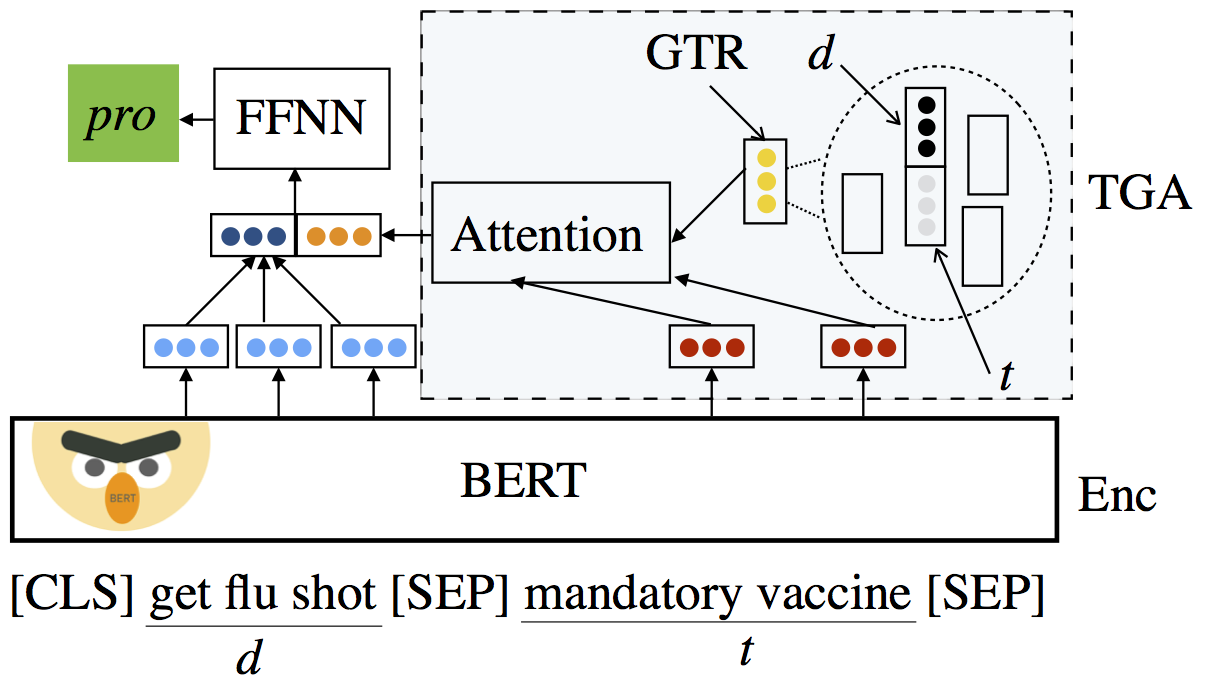}
    \caption{Architecture of TGA Net. Enc indicates contextual conditional encoding (\S \ref{sec:enc}), GTR indicates Generalized Topic Representation (\S \ref{sec:gentop}), TGA indicates Topic-grouped Attention (\ref{sec:tga}).}
    \label{fig:arch}
\end{figure}
\section{Methods}
We develop \textbf{T}opic-\textbf{G}rouped \textbf{A}ttention (TGA) Net: a model to implicitly construct and use relationships between the training and evaluation topics without supervision. The model consists of a contextual conditional encoding layer (\S \ref{sec:enc}), followed by topic-grouped attention (\S \ref{sec:tga}) using generalized topic representations (\S \ref{sec:gentop}) and a feed-forward neural network (see Figure \ref{fig:arch}).

\subsection{Definitions}
Let $D = \{x_i=(d_i, t_i, y_i)\}_{i=1}^N$ be a dataset with $N$ examples, each consisting of a document $d_i$ (a comment), a topic $t_i$, and a stance label $y_i$. Recall that for each unique document $d$, the data may contain
examples with different topics. For example (1) and (2) (Table \ref{tab:dataex}) have the same document but different topics. The task is to predict a stance label $\Hat{y} \in \{\text{pro, con, neutral}\}$ for each $x_i$, based on the \textit{topic-phrase} definition of stance (see \S\ref{sec:related}). 

\subsection{Contextual Conditional Encoding}
\label{sec:enc}
Since computing the stance of a document is dependent on the topic,
prior methods for cross-target stance have found that bidirectional conditional encoding (conditioning the document representation on the topic)  provides large improvements~\citep{Augenstein2016StanceDW}. However, prior work used static word embeddings and we want to take advantage of contextual emebddings. Therefore, we embed a 
document and topic jointly using BERT~\citep{Devlin2019BERTPO}.
That is, we 
treat the document and topic as a sentence pair, and obtain two sequences of token embeddings $\Bar{t} = t^{(1)}, \hdots, t^{(m)}$ for the topic $t$ and $\Bar{d} = d^{(1)}, \hdots, d^{(n)}$ for the document $d$. As a result, the text embeddings are implicitly conditioned on the topic, and vice versa. 

\subsection{Generalized Topic Representations (GTR)}
\label{sec:gentop}
For each example $x=(d, t, y)$ in the data, we compute a generalized topic representation $r_{dt}$: the centroid of the nearest cluster to $x$ in euclidean space, after clustering the training data. We use hierarchical clustering on $v_{dt} = [v_d; v_t]$, a representation of the document $d$ and text $t$, to obtain clusters. We use one $v_d \in \mathbb{R}^E$ and one $v_t \in \mathbb{R}^E$ (where $E$ is the embedding dimension) for each unique document $d$ and unique topic $t$.

To obtain $v_d$ and $v_t$, we first embed the document and topic separately using BERT (i.e., [CLS] $<$text$>$ [SEP] and [CLS] $<$topic$>$ [SEP]) then compute a weighted average over the token embeddings $\Bar{d}$ (and similarly $\Bar{t}$). In this way, $v_d$ ($v_t$) is independent of all topics (comments)
and so $v_d$ and $v_t$ can share information across examples. That is, for examples $x_i,x_j,x_k \in D$ we may have that $d_i = d_j$ but $d_j \neq d_k$ and $t_j = t_k$ but $t_i \neq t_j$.
The token embeddings are weighted in $v_d$ by tf-idf, in order to downplay the impact of common content words (e.g., pronouns or adverbs) in the average. In $v_t$, the token embeddings are weighted uniformly.

\subsection{Topic-Grouped Attention}
\label{sec:tga}
We use the generalized topic representation $r_{dt}$ for example $x$ to compute the similarity between $t$ and other topics in the dataset.
Using learned scaled dot-product attention~\citep{VaswaniAttention}, we compute similarity scores $s_i$ and use these to weigh the importance of the current topic tokens $t^{(i)}$, obtaining a representation $c_{dt}$ that captures the relationship between $t$ and related topics and documents. 
That is, we compute
\begin{equation*}
     c_{dt} = \sum_i s_i t^{(i)}, \ s_i = \text{softmax}\Big(\lambda t^{(i)} \cdot (W_a r_{dt})\Big)
\end{equation*}
where $W_a \in \mathbb{R}^{E \times 2E}$ are learned parameters and $\lambda = 1/\sqrt{E}$ is the scaling value.

\subsection{Label Prediction}
\label{sec:labelpred}
To predict the stance label, we combine the output of our topic-grouped attention with the document token embeddings and pass the result through a feed-forward neural network to compute the output probabilities $p \in \mathbb{R}^3$. That is,
\begin{equation*}
    p = \text{softmax}(W_2 (\tanh (W_1 [\Tilde{d};c_{dt}] + b_1) + b_2))
\end{equation*}
where $\Tilde{d} = \frac{1}{n} \sum_i d^{(i)}$ and $W_1 \in \mathbb{R}^{h \times 2E}, W_2 \in \mathbb{R}^{3 \times h}, b_1 \in \mathbb{R}^h, b_2 \in \mathbb{R}^3$ are learned parameters
and $h$ is the hidden size of the network.
We minimize cross-entropy loss. 
\begin{table}[t]
    \centering
    \begin{tabular}{l|c|c|c}
        \hline
        & \textbf{Train} & \textbf{Dev} & \textbf{Test} \\ \hline
        \# Examples & 13477 & 2062 & 3006 \\
        \# Unique Comments & 1845 & 682 & 786\\ \hline
        \# Few-shot Topics & 638 & 114 & 159 \\
        \# Zero-shot Topics & 4003 & 383 & 600 \\
         \hline
    \end{tabular}
    \caption{Data split statistics for VAST.}
    \label{tab:datasplit}
\end{table}

\section{Experiments}
\subsection{Data}
We split VAST such that all examples $x_i=(d_i,t_i,y_i)$ where $d_i = d$, for a particular document $d$, are in exactly one partition. That is, we randomly assign each unique $d$ to one partition of the data. We assign $70\%$ of unique documents to the training set and split the remainder evenly between development and test.
In the development and test sets we only include examples of types \textit{Heur} and \textit{Corr} (we exclude all noisy \textit{List} examples). 

We create separate zero-shot and few-shot development and test sets. The zero-shot development and test sets consist of topics (and documents) that are not in the training set. 
The few-shot development and test sets consist of topics in the training set (see Table \ref{tab:datasplit}). 
For example, there are $600$ unique topics in the zero-shot test set (\textit{none} of which are in the training set) and $159$ unique topics in the few-shot test set (which \textit{are} in the training set). 
This design ensures that there is no overlap of topics between the training set and the zero-shot development and test sets both for pro/con and neutral examples. We preprocess the data by tokenizing and removing stopwords and punctuation using NLTK.

Due to the linguistic variation in the topic expressions (\S \ref{sec:datanalysis}), we examine the prevalence of lexically similar topics, \textit{LexSimTopics}, (e.g., `taxation policy' vs. `tax policy') between the training and zero-shot test sets. Specifically, we represent each topic in the zero-shot test set and the training set using pre-trained GloVe~\citep{pennington2014glove} word embeddings. Then, test topic $t_i^{(t)}$ is a \textit{LexSimTopic} if there is at least one training topic $t_j^{(r)}$ such that $\text{cosine}\_\text{sim}(t_i^{(t)}, t_j^{(r)}) \geq \theta$ for fixed $\theta \in \mathbb{R}$. 
We manually examine a random sample of zero-shot dev topics to determine an appropriate threshold $\theta$. Using the manually determined threshold $\theta =0.9$, we find that only $16\%$ ($96$ unique topics) of the topics in the entire zero-shot test set are \textit{LexSimTopics}.  

\subsection{Baselines and Models}
We experiment with the following models:
\begin{itemize}
\setlength\itemsep{0em}
    \item \texttt{CMaj}: the majority class computed from each cluster in the training data.
    \item \texttt{BoWV}: we construct separate BoW vectors for the text and topic and pass their concatenation to a logistic regression classifier.
    \item \texttt{C-FFNN}: a feed-forward network trained on the generalized topic representations.
    \item \texttt{BiCond}: a model for cross-target stance that uses bidirectional encoding, whereby the topic is encoded using a BiLSTM as $h_t$ and the text is then encoded using a second BiLSTM conditioned on $h_t$~\citep{Augenstein2016StanceDW}. This model uses fixed pre-trained word embeddings. A weakly supervised version of BiCond is currently state-of-the-art on cross-target TwitterStance.
    \item \texttt{CrossNet}: a model for cross-target stance that encodes the text and topic using the same bidirectional encoding as BiCond and adds an aspect-specific attention layer before classification~\citep{Xu2018CrossTargetSC}. Cross-Net improves over BiCond in many cross-target settings. 
    \item \texttt{BERT-sep}: encodes the text and topic separately, using BERT, and then classification with a two-layer feed-forward neural network.
    \item \texttt{BERT-joint}: contextual conditional encoding followed by a two-layer feed-forward neural network.
    \item \texttt{TGA Net}: our model using contextual conditional encoding and topic-grouped attention.
\end{itemize}
\subsubsection{Hyperparameters}
We tune all models using uniform hyperparameter sampling on the development set. All models are optimized using Adam~\citep{Kingma2015AdamAM}, maximum text length of $200$ tokens (since $<5\%$ of documents are longer) and maximum topic length of $5$ tokens. Excess tokens are discarded

For \texttt{BoWV} we use all topic words and a comment vocabulary of $10,000$ words. We optimize using using L-BFGS and L2 penalty. For \texttt{BiCond} and \texttt{Cross-Net} we use fixed pre-trained $100$ dimensional GloVe \citep{pennington2014glove} embeddings and train for $50$ epochs with early stopping on the development set. For BERT-based models, we fix BERT, train for $20$ epochs with early stopping and use a learning rate of $0.001$. We include complete hyperparameter information in Appendix \ref{app:hyp}. 

We cluster generalized topic representations using Ward hierarchical clustering~\citep{Ward1963HierarchicalGT}, which minimizes the sum of squared distances within a cluster while allowing for variable sized clusters. To select the optimal number of clusters $k$, we randomly sample $20$ values for $k$ in the range $[50,300]$ and minimize the sum of squared distances for cluster assignments in the development set. We select $197$ as the optimal $k$.

\begin{table*}[t]
    \centering
    \begin{tabular}{l|lll|lll|lll}
     & \multicolumn{3}{c|}{F1 All} & \multicolumn{3}{c|}{F1 Zero-Shot} & \multicolumn{3}{c}{F1 Few-Shot} \\ \cline{2-10} 
     & pro & con & all & pro & con & all & pro & con & all \\ \hline
    \texttt{CMaj} & .382 & .441 &  .274&  .389 &  .469 &  .286 &  .375 &  .413 &  .263 \\
    \texttt{BoWV} & .457  &  .402& .372 &  .429 &  .409 &  .349 & .486  & .395 & .393 \\ 
    \texttt{C-FFNN} & .410 & .434 & .300 & .408 & .463 & .417 & .413 & .405  & .282 \\
    \hline
    \texttt{BiCond} & .469 &  .470 &  .415 &  .446 &  .474 & .428 & .489 & .466 &  .400\\
    \texttt{Cross-Net} & .486 & .471  & .455 & .462 & .434 & .434 & .508 & .505 & .474 \\ \hline
    \texttt{BERT-sep} & .4734 & .522 & .5014 & .414 & .506 & .454 & .524 & .539 & .544 \\
    \texttt{BERT-joint} & .545 & \textbf{.591}   &  .653 & .546  & .584  &  .661 & .544   &  \textbf{.597} & .646 \\
    \texttt{TGA Net} & \textbf{.573*} & .590  &  \textbf{.665} &  \textbf{.554} & \textbf{.585} & \textbf{.666}  &  \textbf{.589*}& .595 & \textbf{.663} \\ \hline
    \end{tabular}
\caption{Macro-averaged F1 on the test set. $^*$ indicates significance of \texttt{TGA Net} over \texttt{BERT-joint}, $p<0.05$.}
    \label{tab:results}
\end{table*}
\subsection{Results}
\begin{table*}[t]
\begin{tabular}{ll}
\hline
\textbf{Test Topic} & \textbf{Cluster Topics} \\
\hline
drug addicts & \begin{tabular}[t]{@{}l@{}}war drug, cannabis, legalization, marijuana popularity,  social effect, pot, colorado, \\american lower class, gateway drug, addiction, smoking marijauana, social drug \end{tabular} \\ \hline
oil drilling & \begin{tabular}[t]{@{}l@{}}natural resource, international cooperation, renewable energy, alternative energy,\\ petroleum age, electric car, solar use, offshore drilling, offshore exploration, planet\end{tabular} \\ \hline
\begin{tabular}[t]{@{}l@{}}free college\\ education\end{tabular} & \begin{tabular}[t]{@{}l@{}}tax break home schooling, public school system, education tax, funding education,\\ public service, school tax, homeschool tax credit, community, home schooling parent\end{tabular}\\
\hline
\end{tabular}
\caption{Topics from test examples and training examples in their assigned cluster.}
\label{tab:clusterex}
\end{table*}
We evaluate our models using macro-average F1 calculated on three subsets of VAST (see Table~\ref{tab:results}): all topics, topics only in the test data (zero-shot), and topics in the train or development sets (few-shot). We do this because we want models that perform well on both zero-shot topics and training/development topics. 

We first observe that \texttt{CMaj} and \texttt{BoWV} are strong baselines for zero-shot topics. Next, we observe that \texttt{BiCond} and \texttt{Cross-Net} both perform poorly on our data. 
Although these were designed for cross-target stance, a more limited version of zero-shot stance, they suffer in their ability to generalize across a large number of targets when few examples are available for each. 

We see that while \texttt{TGA Net} and \texttt{BERT-joint} are statistically indistinguishable on all topics, the topic-grouped attention provides a statistically significant improvement for few-shot learning on `pro' examples (with $p < 0.05$). 
Note that conditional encoding is a crucial part of the model, as this provides a large improvement over embedding the comment and topic separately (\texttt{BERT-sep}).
\begin{figure}[t]
    \centering
    \vspace{-10pt}
    \includegraphics[width=.48\textwidth]{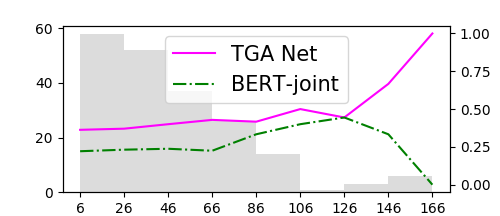}
    \caption{Percentage (right y-axis) each model is best on the test set as a function of the number of \textit{unique topics in each cluster}. Histogram (left y-axis) of unique topics shown in gray.}
    \label{fig:topicpercluster}
    \vspace{-10pt}
\end{figure}
\begin{figure}[t]
    \centering
    \vspace{-10pt}
    \includegraphics[width=.48\textwidth]{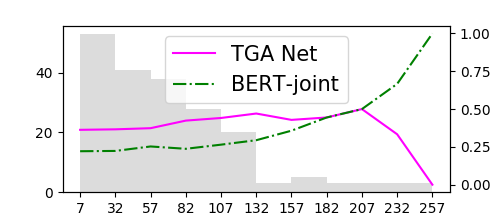}
    \caption{Percentage (right y-axis) each model is best on the test set as a function of the number of \textit{examples per cluster}. Histogram of cluster sizes (left y-axis) shown in gray.}
    \label{fig:expercluster}
    \vspace{-10pt}
\end{figure}

Additionally, we compare the performance of \texttt{TGA Net} and \texttt{BERT-joint} on both zero-shot \textit{LexSimTopics} and non-\textit{LexSimTopics}. We find that while both models exhibit higher performance on zero-shot \textit{LexSimTopics} ($.70$ and $.72$ F1 respectively), these topics are such a small fraction of the zero-shot test topics that zero-shot evaluation primarily reflects model performance on the non-\textit{LexSimTopics}. Additionally, the difference between performance on zero-shot \textit{LexSimTopics} and non-\textit{LexSimTopics} is less for \texttt{TGA Net} (only $0.04$ F1) than for \texttt{BERT-joint} ($0.06$ F1), showing our model is better able to generalize to lexically distinct topics. 

To better understand the effect of topic-grouped attention, we examine the clusters generated in \S\ref{sec:gentop} (see Table~\ref{tab:clusterex}). The clusters range in size from $7$ to $257$ examples (median $62$) with the number of unique topics per cluster ranging from $6$ to $166$ (median $43$). We see that the generalized representations are able to capture relationships between zero-shot test topics and training topics. 


\begin{table}[t]
\begin{tabular}{lllllll}
\hline
 &  & \textit{Imp} &\textit{ mlT} &\textit{ mlS} & \textit{Qte} & \textit{Sarc} \\ \hline
\multicolumn{1}{l|}{\multirow{2}{*}{\begin{tabular}[c]{@{}l@{}}BERT\\ joint\end{tabular}}} & \multicolumn{1}{l|}{I} & .600 & .610 & .541 & .625 & .587 \\
\multicolumn{1}{l|}{} & \multicolumn{1}{l|}{O} & .710 & .748  & .713 & .657 & .662 \\ \hline
\multicolumn{1}{l|}{\multirow{2}{*}{\begin{tabular}[c]{@{}l@{}}TGA \\ Net\end{tabular}}}  & \multicolumn{1}{l|}{I} & .623 & .624 & .547 & .661 & .637 \\
\multicolumn{1}{l|}{} & \multicolumn{1}{l|}{O} & .713 &  .752 & .725 & .663 & .667 \\ \hline
\end{tabular}
\caption{Accuracy on varying phenomena in the test set. I indicates examples with the phenomenon, O indicates examples without.}
\label{tab:lingres}
\end{table}


\begin{table*}[ht]
\centering
\begin{tabular}{lll}
\hline
\textbf{Comment} & \textbf{Topic} & \textbf{$\mathcal{\ell}$}\\ \hline
\begin{tabular}[c]{@{}l@{}}... we \highlight[black]{red!30}{\textbf{\textit{need(-)}}} to get those GOP members out of the House \& Senate,\\ since they only \sout{\textbf{\textit{support(+)}}}$\rightarrow$\highlight[black]{red!30}{\textit{\textbf{patronize(-)}}} billionaire tax breaks,\\ \sout{\textbf{\textit{evidently(+)}}}$\rightarrow$\highlight[black]{red!30}{\textbf{\textit{obviously(-)}}}. We \highlight[black]{red!30}{\textit{\textbf{need(-)}}} MORE PARKS. And they\\should all be \sout{\textbf{\textit{FREE(+)}}}$\rightarrow$\highlight[black]{red!30}{\textbf{\textit{gratuitous(-)}}} ...\end{tabular} & \begin{tabular}[c]{@{}l@{}} government\\spending on\\parks\end{tabular} & Pro\\ \hline

\begin{tabular}[c]{@{}l@{}}... debaters don't \sout{\textbf{\textit{strike(-)}}}$\rightarrow$\highlight[black]{green!30}{\textit{\textbf{shine(+)}}} me as being anywhere near\\diverse in their perspectives on guns. Not one of the gun-gang cited\\any example of where a student with a gun saved someone from\\ something \sout{\textbf{\textit{terrible(-)}}} $\rightarrow$\highlight[black]{green!30}{\textit{\textbf{tremendous(+)}}} on their campuses. At\\ \highlight[black]{red!30}{\textbf{\textit{least(-)}}} the professor speaks up for \highlight[black]{green!30}{\textit{\textbf{rationality(+)}}}.\end{tabular} & guns & Con \\ \hline
\end{tabular}
\caption{Examples with changed majority sentiment polarity. Sentiment words are \textit{\textbf{bold italicized}}, for removed words (\sout{struck out}) and positive (\highlight[black]{green!30}{green (+)}) and negative (\highlight[black]{red!30}{red (-)}) sentiment words. }
\label{tab:sentex}
\end{table*}
\begin{table}[t]
\begin{tabular}{ll|ll}
\hline
 &  & \begin{tabular}[c]{@{}l@{}}\texttt{BERT}\\\texttt{joint}\end{tabular} & \begin{tabular}[c]{@{}l@{}}\texttt{TGA}\\\texttt{Net}\end{tabular} \\ \hline
\multirow{4}{*}{Pro} & $M+$ & .73 & .77 \\
 & $M-$ & .65 & .68 \\ \cline{2-4}
 & $M+ \rightarrow M-$ ($\downarrow$) & .71$\rightarrow$.69 & .74$\rightarrow$.67 \\
 & $M- \rightarrow M+$ ($\uparrow$) & .71$\rightarrow$.74 & .71$\rightarrow$.70 \\ \hline \hline
\multirow{4}{*}{Con} & $M+$ & .74 & .70  \\ 
 & $M-$ & .79  & .80  \\  \cline{2-4}
 & $M+ \rightarrow M-$ ($\uparrow$) & .76$\rightarrow$.80 & .70$\rightarrow$.74  \\
 & $M- \rightarrow M+$ ($\downarrow$) & .75$\rightarrow$.71 & .75$\rightarrow$.74 \\ \hline
\end{tabular}
\caption{F1 on the test set for examples with a majority sentiment polarity ($M$) and conversion between sentiment polarities (e.g., $M+\rightarrow M-$). The direction the score a sentiment-susceptible model is expected to change is indicated with $\uparrow$ or  $\downarrow$. }
\label{tab:sent}
\vspace{-10pt}
\end{table}

We also evaluate the percentage of times each of our best performing models (\texttt{BERT-joint} and \texttt{TGA Net}) is the best performing model on a cluster as a function of the number of unique topics (Figure~\ref{fig:topicpercluster}) and cluster size (Figure~\ref{fig:expercluster}). To smooth outliers, we first bin the cluster statistic and calculate each percent for clusters with at least that value (e.g., clusters with at least $82$ examples). We see that as the number of topics per cluster increases, \texttt{TGA Net} increasingly outperforms \texttt{BERT-joint}. This shows that the model is able to benefit from diverse numbers of topics being represented in the same manner. 
On the other hand, when the number of examples per cluster becomes too large ($>182$), \texttt{TGA NET}'s performance suffers. This suggests that when cluster size is very large, the stance signal within a cluster becomes too diverse for topic-grouped attention to use.

\subsection{Error Analysis}
\subsubsection{Challenging Phenomena}
\label{sec:hard}
We examine the performance of \texttt{TGA Net} and \texttt{BERT-joint} on five challenging phenomena in the data: \textbf{i)}  
\textit{Imp} -- the topic phrase is not contained in the document and the label is not neutral ($1231$ cases), \textbf{ii)} \textit{mlT} -- a document is in examples with multiple topics ($1802$ cases), \textbf{iii)} \textit{mlS} -- a document is in examples with different, non-neutral, stance labels (as in (3) and (4) Table \ref{tab:dataex}) ($952$ cases), \textbf{iv)} \textit{Qte} -- a document with quotations, and \textbf{v)} \textit{Sarc} -- sarcasm, as annotated by~\citet{Habernal2018TheAR}. 

We choose these phenomena to cover a range of challenges for the model. First, \textit{Imp} examples require the model to recognize concepts related to the unmentioned topic in the document (e.g., recognizing that computers are related to the topic `3d printing'). Second, to do well on \textit{mlT} and \textit{mlS} examples, the model must learn more than global topic-to-stance or document-to-stance patterns (e.g., it cannot predict a single stance label for all examples with a particular document). Finally, quotes are challenging because they may repeat text with the opposite stance to what the author expresses themselves (see Appendix Table \ref{tab:apphardex} for examples).

Overall, we find the \texttt{TGA Net} performs better on these difficult phenomena (see Table \ref{tab:lingres}). These phenomena are challenging for both models, as
indicated by the generally lower performance on examples with the phenomena compared to those without, with the \textit{mlS} especially difficult. We observe that \texttt{TGA Net} has particularly large improvements on the rhetorical devices (\textit{Qte} and \textit{Sarc}), suggesting that topic-grouped attention allows the model to learn more complex semantic information in the documents.  

\subsubsection{Stance and Sentiment}
Finally, we investigate the connection between stance and sentiment vocabulary. Specifically, we use the MPQA sentiment lexicon~\citep{Wilson2017MPQAOC} to identify positive and negative sentiment words in texts. We observe that in the test set, the majority ($80\%$) of pro examples have more positive than negative sentiment words, while only $41\%$ of con examples have more negative than positive sentiment words. That is, con stance is often expressed using positive sentiment words but pro stance is rarely expressed using negative sentiment words and therefore there is not a direct mapping between sentiment and stance. 

We use $M+$ to denote majority positive sentiment polarity and similarity for $M-$ and negative. 
We find that on pro examples with $M-$, \texttt{TGA Net} outperforms \texttt{BERT-joint}, while the
reverse is true for con examples with $M+$. For both stance labels and models, performance increases when the majority sentiment polarity agrees with the stance label ($M+$ for pro, $M-$ for con). Therefore, we investigate how susceptible both models are to changes in sentiment. 

To test model susceptibility to sentiment polarity, we generate swapped examples. For examples with majority polarity $p$, we randomly replace sentiment words with a WordNet\footnote{\url{wordnet.princeton.edu}} synonym of opposite polarity until the majority polarity for the example is $-p$ (see Table \ref{tab:sentex}). We then evaluate our models on the examples before and after the replacements. 

When examples are changed from the opposite polarity ($-\rightarrow+$ for pro, $+\rightarrow-$ for con), a model that relies too heavily on sentiment should increase performance. Conversely, when converting \textit{to} the opposite polarity ($+\rightarrow-$ for pro, $-\rightarrow+$ for con) an overly reliant model's performance should decrease. Although the examples contain noise, we find that both models are reliant on sentiment cues, particularly 
when
adding negative sentiment words to a pro stance text. This suggests the models are learning strong associations between negative sentiment and con stance.

Our results also show \texttt{TGA Net} is less susceptible to replacements than \texttt{BERT-joint}. 
On pro $- \rightarrow +$, performance actually decreases by one point (\texttt{BERT-joint} increases by three points) and on con $- \rightarrow +$ performance only decreases by one point (compared to four for \texttt{BERT-joint}). 
\texttt{TGA Net} is better able to distinguish when positive sentiment words are actually indicative of a pro stance, which may contribute to its significantly higher performance on pro.
Overall, \texttt{TGA Net} relies less on sentiment cues than other models.

\section{Conclusion}
We find that our model \texttt{TGA Net}, which uses generalized topic representations to implicitly capture relationships between topics, performs significantly better than BERT for stance detection on pro labels, and performs similarly on other labels. In addition, extensive analysis shows our model provides substantial improvement on a number of challenging phenomena (e.g., sarcasm) and is less reliant on sentiment cues that tend to mislead the models. Our models are evaluated on a new dataset, VAST, that has a large number of topics with wide linguistic variation and that we create and make available. 

In future work we plan to investigate additional methods to represent and use generalized topic information, such as topic modeling. In addition, we will study more explicitly how to decouple stance models from sentiment, and how to improve performance further on difficult phenomena. 

\section{Acknowledgements}
We thank the Columbia NLP group and the anonymous reviewers for their comments. This work is based on research sponsored by DARPA under agreement number FA8750-18-
2-0014. The U.S. Government is authorized to reproduce and distribute reprints for Governmental purposes notwithstanding any copyright notation thereon. The views and conclusions contained herein are those of the authors and should not be interpreted as necessarily representing the official policies or endorsements, either expressed or implied, of DARPA or the U.S. Government.

\bibliography{anthology,emnlp2020,refs}
\bibliographystyle{acl_natbib}

\newpage
\clearpage
\appendix
\section{Appendices}
\subsubsection{Crowdsourcing}
\label{app:crowd}
\begin{figure*}[t]
    \centering
    \includegraphics[width=1.\textwidth]{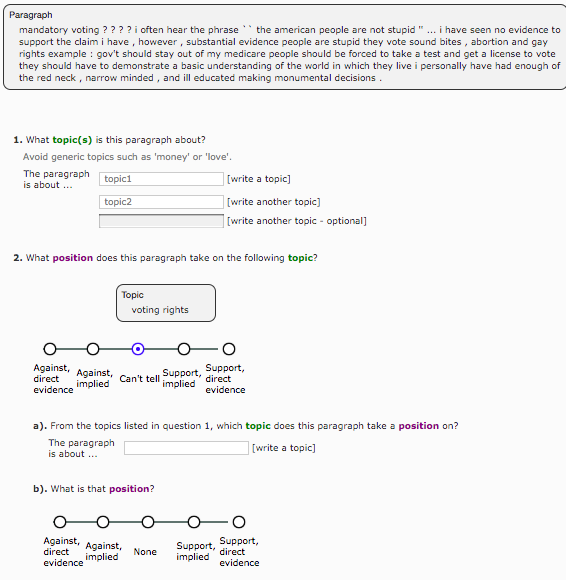}
    \caption{Snapshot of Amazon Mechanical Turk Annotation Task with sample input data.}
    \label{fig:amt}
\end{figure*}

We show a snap shot of one `HIT' of the data annotation task in Figure \ref{fig:amt}. We paid annotators $\$0.13$ per HIT. We had a total of $696$, of which we removed $183$ as a result of quality control. 

\subsubsection{Data}
\label{app:data}
We show complete examples from the dataset in Table \ref{tab:appdataex}. These show the topics extracted from the original ARC stance position, potential annotations and corrections, and the topics listed by annotators as relevant to each comment.

In (a), (d), (i), (j), and (l) the topic make sense to take a position on (based on the comment) and annotators do not correct the topics and provide a stance label for that topic directly. In contrast, the annotators correct the provided topics in (b), (c), (e), (f), (g), (h), and (k). The corrections are because the topic is not possible to take a position on (e.g., `trouble'), or not specific enough (e.g, `california', `a tax break'). In one instance, we can see that one annotator chose to correct the topic (k) whereas another annotator for the same topic and comment chose not to (j). This shows how complex the process of stance annotation is.

We also can see from the examples the variations in how similar topics are expressed (e.g., `public education' vs. `public schools') and the relationship between the stance label assigned for the extracted (or corrected topic) and the listed topic. In most instances, the same label applies to the listed topics. However, we show two instances where this is not the case: (d) -- the comment actually supports `public schools' and (i) -- the comment is actually against `airline'). This shows that this type of example (ListTopic, see \S\ref{sec:crowd}), although somewhat noisy, is generally correctly labeled using the provided annotations. 

We also show neutral examples from the dataset in Table \ref{tab:appneuex}. Examples 1 and 2 were constructed using the process described in \S\ref{sec:neu}. We can see that the new topics are distinct from the semantic content of the comment. Example 3 shows an annotator provided neutral label since the comment is neither in support of or against the topic `women's colleges'. This type of neutral example is less common than the other (in 1 and 2) and is harder, since the comment \textit{is} semantically related to the topic. 

\subsection{Experiments}
\subsubsection{Hyperparameters}
\label{app:hyp}
All neural models are implemented in Pytorch\footnote{\url{https://pytorch.org/}} and tuned on the developement. Our logistic regression model is implemented with scikit-learn\footnote{\url{https://scikit-learn.org/stable/}}. The number of trials and training time are shown in Table \ref{tab:hpsearch}. Hyperparameters are selected through uniform sampling. We also show the hyperparameter search space and best configuration for \texttt{C-FFNN} (Table \ref{tab:hpcffnn}), \texttt{BiCond} (Table \ref{tab:hpbicond}), \texttt{Cross-Net} (Table \ref{tab:hpctsan}), \texttt{BERT-sep} (Table \ref{tab:hpsbert}), \texttt{BERT-joint} (Table \ref{tab:hpjbert}) and \texttt{TGA Net} (Table \ref{tab:hpatt}).  We use one TITAN Xp GPU.  

We calculate expected validation performance~\citep{Dodge2019ShowYW} for F1 in all three cases and additionally show the performance of the best model on the development set (Tale \ref{tab:expval}). Models are tuned on the development set and we use macro-averaged F1 of all classes for zero-shot examples to select the best hyperparameter configuration for each model. We use the scikit-learn implementation of F1. We see that the improvement of \texttt{TGA Net} over \texttt{BERT-joint} is high on the development set.

\subsubsection{Results}
\subsubsection{Error Analysis}
We investigate the performance of our two best models (\texttt{TGA Net} and \texttt{BERT-joint}) on five challenging phenomena, as discussed in \S\ref{sec:hard}. The phenomena are:
\begin{itemize}
\setlength\itemsep{0em}
    \item Implicit (\textit{Imp}): the topic is not contained in the document.
    \item Multiple Topics (\textit{mlT}): document has more than one topic.
    \item Multiple Stance (\textit{mlS}):  a document has examples with different, non-neutral, stance labels.
    \item Quote (\textit{Qte}): the document contains quotations.
    \item Sarcasm (\textit{Sarc}): the document contains sarcasm, as annotated by \citet{Habernal2018TheAR}.
\end{itemize}
We show examples of each of these phenomena in Table \ref{tab:apphardex}. 

\subsubsection{Stance and Sentiment}
To construct examples with swapped sentiment words we use the MPQA lexicon~\citep{Wilson2017MPQAOC} for sentiment words. We use WordNet to select synonyms with opposite polarity, ignoring word sense and part of speech. We show examples from each set type of swap, $+ \rightarrow -$ (Table \ref{tab:pos2negwords}) and $- \rightarrow +$ (Table \ref{tab:neg2poswords}). In total there are $1158$ positive sentiment words and $1727$ negative sentiment words from the lexicon in our data. Of these, $218$ positive words have synonyms with negative sentiment, and $224$ negative words have synonyms with positive sentiment. 
\begin{table*}[h]
    \centering
    \begin{tabular}{lcllc|c}
    \hline
        \textbf{Comment} & \begin{tabular}[t]{@{}l@{}}\textbf{ARC}\\\textbf{Stance}\end{tabular} & \begin{tabular}[t]{@{}l@{}}\textbf{Extracted}\\\textbf{Topic}\end{tabular} & \begin{tabular}[t]{@{}l@{}}\textbf{Listed}\\\textbf{Topic}\end{tabular} & \textbf{$\mathcal{\ell}$}\\
         \hline
\multirow{5}{*}{\begin{tabular}[t]{@{}l@{}}  So based on these numbers \highlight[black]{yellow!30}{London}\\\highlight[black]{yellow!30}{is forking out ~12-24 Billion dollars}\\\highlight[black]{yellow!30}{to pay for the Olympics}. According\\to the Official projection they have\\already spent 12 Billion pounds (or\\just abous \$20 billion). Unofficially\\the bill is looking more like 24 Billion\\pounds (or closer to 40 Billion dollars).\\ What a \highlight[black]{yellow!30}{complete waste of Money}. \end{tabular}} & 
\multirow{5}{*}{\begin{tabular}[t]{@{}l@{}} \highlight[black]{green!30}{\textit{Olympics}}\\are more\\\highlight[black]{green!30}{\textit{trouble}} \end{tabular}} & \multirow{2}{*}{\textbullet olympics} & \textbullet olympics & C & a\\

& & & \begin{tabular}[t]{@{}l@{}}\textbullet london \\olympics\\budget\end{tabular} & & \\ \cdashline{3-6}

& & \begin{tabular}[t]{@{}l@{}} \sout{trouble}\rotatebox[origin=c]{270}{$\Rsh$} \\ \textbullet spending \end{tabular} &
\multirow{2}{*}{\begin{tabular}[t]{@{}l@{}} \\\textbullet wasting \\money\end{tabular}} & C & b\\

& & \begin{tabular}[t]{@{}l@{}}money on\\the olympics \end{tabular} & \multirow{2}{*}{\textbullet sport} & & \\ \cdashline{3-6}

& & \begin{tabular}[t]{@{}l@{}} \sout{trouble}\rotatebox[origin=c]{270}{$\Rsh$} \\ \textbullet olympics \end{tabular} & \textbullet \begin{tabular}[t]{@{}l@{}}london\\finances\end{tabular} & C & c \\
\hline
\multirow{3}{*}{\begin{tabular}[t]{@{}l@{}} The era when there were no public\\schools was not a good socio-economic\\time in the life of our nation. Anything\\which weakens \highlight[black]{yellow!30}{public schools and their}\\\highlight[black]{yellow!30}{funding} will result in the most\\vulnerable \highlight[black]{yellow!30}{youth of America} being left\\out of the chance to get an education. \end{tabular}} &  \multirow{3}{*}{\begin{tabular}[t]{@{}l@{}} \highlight[black]{green!30}{\textit{Home}}\\\highlight[black]{green!30}{\textit{schoolers}}\\do not\\deserve a\\\highlight[black]{green!30}{\textit{tax break}} \end{tabular}} & \begin{tabular}[t]{@{}l@{}}\textbullet home\\schoolers\end{tabular} & \textbullet \begin{tabular}[t]{@{}l@{}}public\\schools \end{tabular} & C  & d\\  \cdashline{3-6}

&  & \begin{tabular}[t]{@{}l@{}} \sout{a tax break}\rotatebox[origin=c]{270}{$\Rsh$} \\ \textbullet public schools \end{tabular} & \begin{tabular}[t]{@{}l@{}} \\ \textbullet youth of\\america\end{tabular} & P & e\\ \cdashline{3-6}

& & \begin{tabular}[t]{@{}l@{}} \sout{a tax break}\rotatebox[origin=c]{270}{$\Rsh$} \\\textbullet public education \end{tabular} & \begin{tabular}[t]{@{}l@{}} \\\textbullet public\\education\\ funding\end{tabular} & P & f\\ \hline
\multirow{3}{*}{\begin{tabular}[t]{@{}l@{}}\highlight[black]{yellow!30}{Airports and the roads} on east nor west\\coast can not handle the present volume\\adequately as is. I did ride the vast\\\highlight[black]{yellow!30}{trains} in Europe, Japan and China and\\found them very comfortable and\\providing much better connections and\\more efficient.\end{tabular}} & \multirow{3}{*}{\begin{tabular}[t]{@{}l@{}} \highlight[black]{green!30}{\textit{California}}\\needs\\\highlight[black]{green!30}{\textit{high-speed}}\\\highlight[black]{green!30}{\textit{rail}}\end{tabular}} & \begin{tabular}[t]{@{}l@{}} \sout{california}\rotatebox[origin=c]{270}{$\Rsh$} \\ \textbullet train\end{tabular} & \begin{tabular}[t]{@{}l@{}}\\\textbullet transportation\end{tabular} & P & g\\ \cdashline{3-6}

& & \begin{tabular}[t]{@{}l@{}} \sout{california}\rotatebox[origin=c]{270}{$\Rsh$} \\\textbullet traffic\end{tabular} & \begin{tabular}[t]{@{}l@{}}\\\textbullet roadway\end{tabular} & C & h \\ \cdashline{3-6}

& & \begin{tabular}[t]{@{}l@{}}\textbullet high-speed\\rail\end{tabular} &  \begin{tabular}[t]{@{}l@{}}\textbullet airline\\\textbullet public\\transit\end{tabular} & P & i\\ \hline
\multirow{3}{*}{\begin{tabular}[t]{@{}l@{}} There is only a shortage of \highlight[black]{yellow!30}{agricultural}\\\highlight[black]{yellow!30}{labor} at \highlight[black]{yellow!30}{current wages}. Raise the wage\\to a fair one, and  \highlight[black]{yellow!30}{legal workers} will do\\it. If US agriculture is unsustainable\\without abusive labor practices, should\\we continue to prop up those practices? \end{tabular}} & \multirow{3}{*}{\begin{tabular}[t]{@{}l@{}} \highlight[black]{green!30}{\textit{Farms}}\\ could\\ survive \\ without \\ \highlight[black]{green!30}{\textit{illegal}}\\\highlight[black]{green!30}{\textit{labor}}\end{tabular}} & \textbullet farms & \textbullet \begin{tabular}[t]{@{}l@{}}agricultural\\labor\end{tabular} & C & j\\ \cdashline{3-6}

& &\begin{tabular}[t]{@{}l@{}}\sout{farms}\rotatebox[origin=c]{270}{$\Rsh$}\\\textbullet illegal workers \end{tabular} &  \begin{tabular}[t]{@{}l@{}} \\\textbullet agricultural\\labor wages\end{tabular} & C & k\\ \cdashline{3-6}

& & \textbullet illegal labor & \begin{tabular}[t]{@{}l@{}}\textbullet agricultural\\labor\\\textbullet labor\end{tabular} & C & l\\ \hline
    \end{tabular}
    \caption{Complete examples from our dataset with extracted topics (\highlight[black]{green!30}{\textit{green, italic}}) and corrections (old \sout{struck out}). Topics related to the comment are also shown (listed topics), as are labels (\textbf{$\mathcal{\ell}$}), where P indicates Pro and C indicates C. Each label applies to all topics in the cell. Phrases related to the corrections or listed topics are highlighted in \highlight[black]{yellow!30}{yellow}.}
    \label{tab:appdataex}
\end{table*}
\begin{table*}[t]
\centering
\begin{tabular}{llll||c}
\hline
    \textbf{Comment} &\begin{tabular}[t]{@{}l@{}}\textbf{Original}\\\textbf{Topic}\end{tabular} &\textbf{$\mathcal{\ell}$} & \begin{tabular}[t]{@{}l@{}}\textbf{New}\\\textbf{Topic}\end{tabular} &  \\  \hline

\begin{tabular}[t]{@{}l@{}}
    Good idea. I have always had a cat or two. While being\\inhumane, declawing places a cat in danger. Should my\\charming indoor kitty somehow escape outside, he would\\have no way to defend himself.Why don't humans have\\their finger-and tonails removed to save on manicures?\\Answer:they are important to the functioning and\\protection of our bodies.\end{tabular} & nail removal & Pro &  attack & 1 \\ \hline
    
\begin{tabular}[t]{@{}l@{}}Marijuana is not addictive -- and is much less dangerous\\than alcohol. The gate-way drugs are prescription meds\\found in medicine cabinets everywhere. Heroin is a lot less\\expensive than marijuana and if marijuana were legal, and\\less expensive, fewer people would want heroin. \end{tabular} & \begin{tabular}[t]{@{}l@{}}prescription\\meds\end{tabular} & Con & israel & 2\\ \hline

\begin{tabular}[t]{@{}l@{}}There are no women only law schools. Womens' colleges,\\like Mills College, that do offer graduate degrees have\\co-ed graduate schools. The example of Hillary Clinton's\\success at Yale Law School either says nothing about\\womens' colleges or supports them.\end{tabular} & \begin{tabular}[t]{@{}l@{}}women's\\colleges\end{tabular} & N & \begin{tabular}[t]{@{}l@{}}women's\\colleges\end{tabular} & 3 \\ \hline
\end{tabular}
\caption{Neutral examples from the dataset. N indicates neutral original label}
\label{tab:appneuex}
\end{table*}
\begin{table*}[h]
    \centering
    \begin{tabular}{l|cccccc}
        \hline
        & TGA Net & BERT-joint & BERT-sep & BiCond & Cross-Net & C-FFNN\\ \hline
        \# search trials & 10 & 10 & 10 & 20 & 20 & 20 \\
        Training time (seconds) & 6550.2 & 2032.2 & 1995.6  & 2268.0 & 2419.8 & 5760.0\\ 
        \# Parameters & 617543 & 435820 & 974820 & 225108 & 205384 & 777703\\
        \hline
    \end{tabular}
    \caption{Search time and trials for various models.}
    \label{tab:hpsearch}
\end{table*}
\begin{table*}[t]
    \centering
    \begin{tabular}{lcc}
        \textbf{Hyperparameter} & \textbf{Search space} & \textbf{Best Assignment}\\
        \hline
        batch size & 64 & 64 \\
        epochs & 50 & 50 \\
        dropout & \textit{uniform-float}[$0.1,0.3$] & $0.28149172466319095$\\
        hidden size & \textit{uniform-integer}[$300,1000$] & $505$\\
        learning rate & $0.001$ & $0.001$ \\
        learning rate optimizer & Adam & Adam \\
        \hline
    \end{tabular}
    \caption{Hyperparameter search space and setting for \texttt{C-FFNN}.}
    \label{tab:hpcffnn}
\end{table*}
\begin{table*}[t]
    \centering
    \begin{tabular}{lcc}
        \textbf{Hyperparameter} & \textbf{Search space} & \textbf{Best Assignment}\\
        \hline
        batch size & 64 & 64\\
        epochs & 100 & 100 \\
        dropout & \textit{uniform-float}[$0.1,0.5$] & $04850254141775727$  \\
        hidden size & \textit{uniform-integer}[$40,100$] & $78$\\
        learning rate & \textit{loguniform}[$0.001,0.01$] & $0.004514020306243207$ \\
        learning rate optimizer & Adam & Adam \\
        pre-trained vectors & Glove & Glove \\
        pre-trained vector dimension & 100 & 100 \\ \hline
    \end{tabular}
    \caption{Hyperparameter search space and setting for \texttt{BiCond}.}
    \label{tab:hpbicond}
\end{table*}
\begin{table*}[t]
    \centering
    \begin{tabular}{lcc}
        \textbf{Hyperparameter} & \textbf{Search space} & \textbf{Best Assignment}\\
        \hline
        batch size & 64 & 64\\
        epochs & 100 & 100 \\
        dropout & \textit{uniform-float}[$0.1,0.5$] & $0.36954545196802335$ \\
        BiLSTM hidden size &  \textit{uniform-integer}[$40,100$] & $68$\\
        linear layer hidden size & \textit{uniform-integer}[$20,60$] & $48$\\
        attention hidden size & \textit{uniform-integer}[$20,100$] & $100$\\
        learning rate & \textit{loguniform}[$0.001,0.01$] & $0.00118168557993075$\\
        learning rate optimizer & Adam & Adam \\
        pre-trained vectors & Glove & Glove \\
        pre-trained vector dimension & 100 & 100 \\ \hline
    \end{tabular}
    \caption{Hyperparameter search space and setting for \texttt{Cross-Net}.}
    \label{tab:hpctsan}
\end{table*}
\begin{table*}[t]
    \centering
    \begin{tabular}{lcc}
        \textbf{Hyperparameter} & \textbf{Search space} & \textbf{Best Assignment}\\
        \hline
        batch size & $64$ & $64$ \\
        epochs & $20$ & $20$ \\
        dropout & \textit{uniform-float}[$0.1,0.3$] & $0.22139772968435562$\\
        hidden size & \textit{uniform-integer}[$300,1000$] & $633$ \\
        learning rate & $0.001$ & $0.001$\\
        learning rate optimizer & Adam & Adam \\ \hline
    \end{tabular}
    \caption{Hyperparameter search space and setting for \texttt{BERT-sep}.}
    \label{tab:hpsbert}
\end{table*}
\begin{table*}[t]
    \centering
    \begin{tabular}{lcc}
        \textbf{Hyperparameter} & \textbf{Search space} & \textbf{Best Assignment}\\
        \hline
        batch size & 64 & 64 \\
        epochs & 20 & 20 \\
        dropout & \textit{uniform-float}[$0.1,0.4$] & $0.20463604390811982$\\
        hidden size & \textit{uniform-integer}[$200,800$] & $283$ \\
        learning rate & $0.001$ & $0.001$ \\
        learning rate optimizer & Adam & Adam \\ 
        \hline
    \end{tabular}
    \caption{Hyperparameter search space and setting for \texttt{BERT-joint}.}
    \label{tab:hpjbert}
\end{table*}
\begin{table*}[t]
    \centering
    \begin{tabular}{lcc}
        \textbf{Hyperparameter} & \textbf{Search space} & \textbf{Best Assignment}\\
        \hline
        batch size & $64$ & $64$ \\
        epochs & $50$ & $50$ \\
        dropout & \textit{uniform-float}[$0.1,0.3$] & $0.35000706311476193$\\
        hidden size & \textit{uniform-integer}[$300,1000$] & $401$ \\
        learning rate & $0.001$ & $0.001$\\
        learning rate optimizer & Adam & Adam \\ \hline
    \end{tabular}
    \caption{Hyperparameter search space and settings for \texttt{TGA Net}.}
    \label{tab:hpatt}
\end{table*}
\begin{table*}[t]
    \centering
    \begin{tabular}{l|ccc|ccc}
    \hline
        & \multicolumn{3}{c|}{Best Dev} & \multicolumn{3}{c}{$\mathbb{E}$[Dev]}\\
         & F1$_a$ & F1$_z$ & F1$_f$ & F1$_a$ & F1$_z$ & F1$_f$ \\ \hline
         CMaj & .3817 & .2504 & .2910 & -- & -- & --\\
         BoWV & .3367 & .3213 & .3493 & -- & -- & -- \\
        C-FFNN & .3307 & .3147 & .3464& .3315 & .3128 & .3590 \\
        BiCond & .4229& .4272& .4170 & .4423 & .4255& .4760\\
        Cross-Net & .4779& .4601& .4942& .4751& .4580 & .4979\\
        BERT-sep & .5314 & .5109 & .5490 & .5308 & .5097  & .5519 \\
        BERT-joint & .6589 & .6375 & .6099 & .6579 & .6573& .6566\\
        TGA Net & .6657 & .6851  & .6421 & .6642 & .6778& .6611 \\ \hline
    \end{tabular}
    \caption{Best results on the development set and expected validation score \citep{Dodge2019ShowYW} for all tuned models. $a$ is All, $z$ is zero-shot, $f$ is few-shot.}
    \label{tab:expval}
\end{table*}
\begin{table*}[h]
    \centering
    \begin{tabular}{c|llc}
        \hline
        \textbf{Type} & \textbf{Comment} & \textbf{Topic} &\textbf{$\mathcal{\ell}$}  \\
         \hline
        \textit{Imp} & \begin{tabular}[t]{@{}l@{}}No, it's not just that the corporations will have larger printers.\\It is that most of us will have various sizes of printers. IT's\\just what happened with computers. I was sold when some\\students from Equador showed me their easy to make, working,\\prosthetic arm. Cost to make, less than one hundred dollars.\end{tabular} & \textbullet 3d printing & Pro \\ \hline
        
        \textit{Sarc}& \begin{tabular}[t]{@{}l@{}}yes, let's hate cyclists: people who get off their ass and ride,\\ staying fit as they get around the city. they don't pollute the air,\\they don't create noise, they don't create street after street\\clogged with cars dripping oil... I think the people who hate\\ cyclists are the same ones who hate dogs: they have tiny little\\shards of coal where their heart once was. they can't move fast\\or laugh, and want no one else to, either. According to the\\DMV, in 2009 there were 75,539 automobile crashes in\\New York City, less than 4 percent of those crashes involved\\a bicycle. cyclists are clearly the problem here.\end{tabular} & \textbullet cyclists & Pro \\ \hline
        
        \textit{Qte} & \begin{tabular}[t]{@{}l@{}}``cunning, baffling and powerful disease of addiction'' - LOL no.\\This is called 'demon possession'. Let people do drugs. They'll\\go through a phase and then they'll get tired of it and then\\ they'll be fine. UNLESS they end up in treatment and must\\confess a disease of free will, in which case all bets are off.\end{tabular} & \textbullet \begin{tabular}[t]{@{}l@{}} disease of\\addiction\end{tabular} & Con \\ \hline 
        
        \textit{mlS} & \begin{tabular}[t]{@{}l@{}}That this is even being debated is evidence of the descent of\\American society into madness. The appalling number of gun\\deaths in America is evidence that more guns would make\\society safer? Only in the US does this kind of logic translate\\into political or legal policy. I guess that's what exceptionalism\\ means.\end{tabular} & \begin{tabular}[t]{@{}l@{}} \textbullet guns\\ \textbullet gun control\end{tabular} & \begin{tabular}[t]{@{}l@{}} Con \\ Pro \end{tabular}\\ \hline
        
        \textit{mlT} & \begin{tabular}[t]{@{}l@{}}The focus on tenure is just another simplistic approach to\\changing our educational system. The judge also overlooked\\that tenure can help attract teachers. Living in West Virginia,\\a state with many small and isolated communities, why would\\any teacher without personal ties to our state come here, if she\\can fired at will? I know that I and my wife would not.\end{tabular} & \begin{tabular}[t]{@{}l@{}}\textbullet tenure \\ \textbullet stability \end{tabular} & \begin{tabular}[t]{@{}l@{}} Pro \\ Pro\end{tabular} \\ \hline
    \end{tabular}
    \caption{Examples of hard phenomena in the dataset as discussed in \S\ref{sec:hard}.}
    \label{tab:apphardex}
\end{table*}
\begin{table}[h]
    \centering
    \begin{tabular}{ll}
        \hline
        \begin{tabular}[t]{@{}l@{}}\highlight[black]{red!30}{\textbf{Negative(-)}}\\\textbf{Word}\end{tabular} & \begin{tabular}[t]{@{}l@{}}\highlight[black]{green!30}{\textbf{Positive(+)}}\\\textbf{Word}\end{tabular}\\\hline
        inevitably & necessarily  \\ 
        low & humble\\
        resistant & tolerant\\
        awful & tremendous\\
        eliminate & obviate\\
        redundant & spare\\
        rid & free\\
        hunger & crave\\
        exposed & open\\
        mad & excited \\
        indifferent & unbiased\\
        denial & defense\\
        costly & dear\\
        weak & light\\
        laughable & amusing\\
        worry & interest\\
        pretend & profess\\
        depression & impression\\
        fight & press\\
        trick & joke\\
        slow & easy\\
        sheer & bold\\
        doom & destine\\
        wild & fantastic\\
        laugh & jest\\
        partisan & enthusiast\\
        deep & rich\\
        restricted & qualified\\
        gamble & adventure\\
        shake& excite\\
        scheme & dodge\\
        suffering & brook\\
        burn & glow\\
        argue & reason\\
        oppose & defend\\ 
        hard & strong\\
        complicated & refine\\
        fell & settle\\
        avoid & obviate\\
        hedge & dodge\\ \hline
        
    \end{tabular}
    \caption{Example word pairs for converting words from \highlight[black]{red!30}{negative(-)} to \highlight[black]{green!30}{positive(+)} sentiment.}
    \label{tab:neg2poswords}
\end{table}
\begin{table}[h]
    \centering
    \begin{tabular}{ll}
        \hline
        \begin{tabular}[t]{@{}l@{}}\highlight[black]{green!30}{\textbf{Positive(+)}}\\\textbf{Word}\end{tabular} & \begin{tabular}[t]{@{}l@{}}\highlight[black]{red!30}{\textbf{Negative(-)}}\\\textbf{Word}\end{tabular}\\\hline
        compassion & pity\\
        terrified & terrorize\\
        frank & blunt\\
        modest & low\\
        magic & illusion\\ 
        sustained & suffer\\
        astounding & staggering\\
        adventure & gamble\\
        glow & burn\\
        spirited & game\\ 
        enduring & suffer\\
        wink & flash\\
        sincere & solemn\\
        amazing & awful\\
        triumph & wallow\\ 
        compassionate & pity\\
        plain & obviously\\
        stimulating & shake\\
        excited & mad\\
        sworn & swear\\ 
        unbiased & indifferent\\
        compelling & compel\\
        exciting & shake\\
        yearn & ache\\
        validity & rigor\\ 
        seasoned & temper\\
        appealing & sympathetic\\
        innocent & devoid\\
        pure & stark\\
        super & extremely\\ 
        interesting & worry\\
        productive & fat\\
        strong & stiff\\
        fortune & hazard\\
        rally & bait\\ 
        motivation & need\\
        ultra & radical\\
        justify & rationalize\\
        amusing & laughable\\
        awe & fear\\ 
 \hline
    \end{tabular}
    \caption{Example word pairs for converting words from \highlight[black]{green!30}{positive(+)} to \highlight[black]{red!30}{negative(-)} sentiment.}
    \label{tab:pos2negwords}
\end{table}

\end{document}